\documentclass[runningheads]{llncs}
\usepackage[T1]{fontenc}
\usepackage{booktabs}
\usepackage{array}
\usepackage{multirow}
\usepackage{graphicx}
\usepackage{marvosym} % 确保已加载宏包
\usepackage{amssymb}
\usepackage{amsmath}

\begin{document}
%
%\title{A Scalable Transformer Method for Multi-Agent Audio-Visual Navigation in 3D Environments}
\title{Advancing Audio-Visual Navigation Through Multi-Agent Collaboration in 3D Environments}
\titlerunning{MASTAVN}
\author{
  \small Hailong Zhang\inst{1} \and
   \small  Yinfeng Yu\inst{1}\textsuperscript{(\Letter)} \and
    \small Liejun Wang\inst{1} \and
    \small Fuchun Sun\inst{2} \and
    \small Wendong Zheng\inst{3}
}
\authorrunning{HLZ, YYF et al.}
\institute{
   \small  Xinjiang Multimodal Intelligent Processing and Information Security Engineering Technology Research Center, \\
    \small School of Computer Science and Technology, Xinjiang University, Urumqi 830017, China \\
  \email{yuyinfeng@xju.edu.cn} \and
   \small  Department of Computer Science and Technology, Tsinghua University, Beijing 100091, China \and
   \small  School of Electrical Engineering and Automation, Tianjin University of Technology, Tianjin 300382, China
}

% 对通讯作者做说明（无编号脚注）
\renewcommand{\thefootnote}{}  
\footnotetext{\textsuperscript{(\Letter)} \small Yinfeng Yu is the corresponding author (e-mail: yuyinfeng@xju.edu.cn).}
\maketitle              % typeset the header of the contribution
\begin{abstract}

Intelligent agents often require collaborative strategies to achieve complex tasks beyond individual capabilities in real-world scenarios. While existing audio-visual navigation (AVN) research mainly focuses on single-agent systems, their limitations emerge in dynamic 3D environments where rapid multi-agent coordination is critical, especially for time-sensitive applications like emergency response. This paper introduces MASTAVN (Multi-Agent Scalable Transformer Audio-Visual Navigation), a scalable framework enabling two agents to collaboratively localize and navigate toward an audio target in shared 3D environments. By integrating cross-agent communication protocols and joint audio-visual fusion mechanisms, MASTAVN enhances spatial reasoning and temporal synchronization. Through rigorous evaluation in photorealistic 3D simulators (Replica and Matterport3D), MASTAVN achieves significant reductions in task completion time and notable improvements in navigation success rates compared to single-agent and non-collaborative baselines. This highlights the essential role of spatiotemporal coordination in multi-agent systems. Our findings validate MASTAVN's effectiveness in time-sensitive emergency scenarios and establish a paradigm for advancing scalable multi-agent embodied intelligence in complex 3D environments.

\end{abstract}
\section{Introduction}

\begin{figure}[h] % 使用标准的浮动位置参数
  \includegraphics[width=\textwidth]{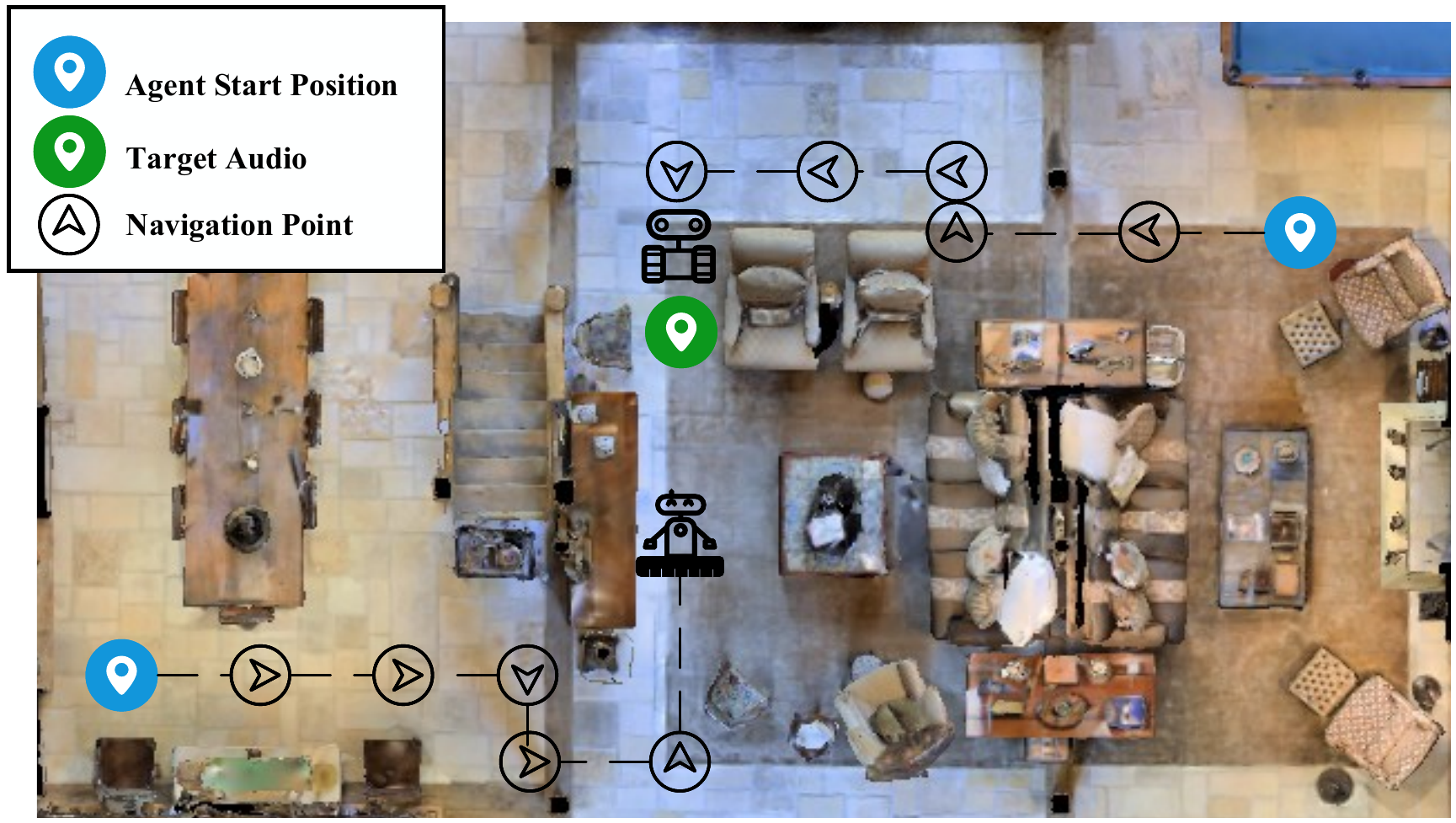}
  \caption{
  Multi-agent audio-visual embodied navigation: multiple intelligent agents approach the target based on the location of the target audio.
  }
  \label{fig1}
\end{figure}

In navigation tasks~\cite{an2024etpnav,chen2022reinforced,jain2019two,paul2022avlen,tampuu2017multiagent,wang2021structured,wu2024vision,fsaavn,yu2023echo,YinfengIJCAI2023MACMA,yu2022sound,yu2025dope,zhu2020vision}, a 3D environment often contains multiple types of robots, and it is possible that a single agent may not be able to perceive audio information or reach an audio target. However, collaboration among multiple agents can effectively compensate for these shortcomings. The advantages of multi-agent systems include: (1) Each agent may be engaged in specific tasks or idle, resulting in different initial positions when facing extraordinary situations, which allows them to cover a larger reachable area compared to single-agent systems; (2) Implicit interactions between agents facilitate the fusion of multimodal information and inter-agent communication. These advantages can significantly improve the success rate of audio-visual navigation tasks while reducing navigation failures in dangerous or urgent scenarios.

Although existing methods~\cite{chen2020soundspaces,chen2021waypoints,chen2022soundspaces,chen2023omnidirectional,huang2023audio,majumder2021move2hear,wei2022learning,yang2024rila} have achieved considerable success in audio-visual navigation and performed well in terms of evaluation metrics, there remain several pressing challenges in real-world applications. In real-world audio-visual navigation tasks, even a few seconds of navigation time difference can lead to drastically different outcomes. Moreover, real environments often involve multiple autonomous agents, raising the question of effectively utilizing multi-agent collaboration. Current audio-visual navigation methods fall short in adequately addressing these urgent issues.

To tackle these challenges, we propose a novel audio-visual navigation framework, MASTAVN (Multi-Agent Scalable Transformer Audio-Visual Navigation). Compared to existing studies, our experimental results demonstrate that this model enables effective multimodal information fusion and deep interaction among multiple agents. It accelerates the learning of decision-making strategies from multimodal inputs. It offers a scalable framework where expanding the number of encoders and the dimensions of decoders can accommodate more agents. As illustrated in Figure~\ref{fig1}, two agents interact implicitly through the MASTAVN framework to accomplish an audio-visual navigation task. In summary, our main contributions are as follows:
\begin{itemize}
\item We developed a dual-agent audio-visual navigation platform to better adapt to practical application environments. Our platform is the first to achieve scalable multi-agent audio-visual navigation compared to previous environments.
\item Experimental validation in real-world 3D scenes from the Replica~\cite{straub2019replica} and Matterport3D~\cite{chang2017matterport3d} datasets shows that the trained dual agents can significantly improve navigation success rates and efficiency through collaboration.
\item We explore a sequence modeling approach for multi-agent collaboration, providing insights and experience that can serve as valuable references for future research in multi-agent audio-visual navigation.
\end{itemize}

\section{Related Work}

Audio-Visual Navigation~\cite{chen2020soundspaces,chen2021waypoints,chen2022soundspaces,chen2023omnidirectional}. As a critical task in embodied navigation~\cite{chen2022soundspaces,chen2025affordances,chen2023omnidirectional,chen2024webvln,long2024discuss,wu2024embodied,zhang2025mapnav,zheng2024towards}, audio-visual navigation requires agents to locate sound sources in real, unmapped 3D environments. Early research~\cite{chen2020soundspaces,chen2021waypoints,chen2022soundspaces,chen2023omnidirectional} on single-agent navigation has thoroughly explored the potential of audio-visual information fusion. Gan et al.~\cite{gan2020look} were the first to systematically propose an audio-visual fusion navigation framework, highlighting the central role of multimodal perception in navigating complex environments. Chen et al.~\cite{chen2020soundspaces} marked a milestone in audio-visual navigation by introducing SoundSpaces, the first platform to simulate embedded audio-visual navigation. They also proposed the AV-WAN~\cite{chen2021waypoints} algorithm, which sets waypoints as subgoals to decompose the navigation process, further enhancing the agent's navigation ability. J. Chen et al.~\cite{chen2023omnidirectional} combined omnidirectional information collection with cross-domain knowledge transfer strategies, significantly improving the generalization capabilities of navigation systems. These studies have collectively advanced the development of audio-visual navigation step by step. Unlike earlier works on single agents, our approach emphasizes deep interaction and parallel processing among multiple agents in complex environments. We highlight the importance of implicit multi-agent interaction in complex 3D scenarios to boost navigation efficiency significantly.

Vision-Based Multi-Agent Reinforcement Learning~\cite{jain2019two,tampuu2017multiagent}. In the field of multi-agent reinforcement learning, various methods have been proposed to address multi-agent problems, spanning settings from cooperative to competitive agents~\cite{jain2019two,tampuu2017multiagent}. Recently, research has increasingly focused on enabling communication and cooperation among multiple agents in rich visual environments. However, most of these studies have not extended into the domain of audio-visual navigation. Our research introduces a scalable multi-agent platform into the audio-visual setting, enabling deep integration and interaction among agents through early implicit communication. This offers new perspectives and methodologies for multi-agent audio-visual navigation tasks.

MAT (Multi-Agent Transformer)~\cite{wen2022multi}, a multi-agent reinforcement learning model based on the Transformer architecture~\cite{vaswani2017attention}, presents an innovative sequential modeling approach for solving multi-agent collaboration problems. This work demonstrates the significant advantages of Transformers over traditional reinforcement learning methods in handling multi-agent interaction tasks. MAT~\cite{wen2022multi} reformulates multi-agent policy optimization as a sequential decision-making process and utilizes an encoder-decoder architecture to model complex inter-agent relationships efficiently, achieving monotonic policy performance improvements. Inspired by MAT~\cite{wen2022multi}'s serialization of the multi-agent decision-making process, we designed a novel collaborative multi-agent architecture, which provides a strong theoretical and methodological foundation for our research, significantly advancing the design and optimization of our model.

\section{Approach}
In this section, we first introduce the four main steps involved in achieving multi-agent audio-visual navigation. Then, we discuss the core MAST module used in this process. Finally, we explain the relationship between Transformer models and multi-agent tasks.

\begin{figure}[h] % 使用标准的浮动位置参数
  \includegraphics[width=\textwidth]{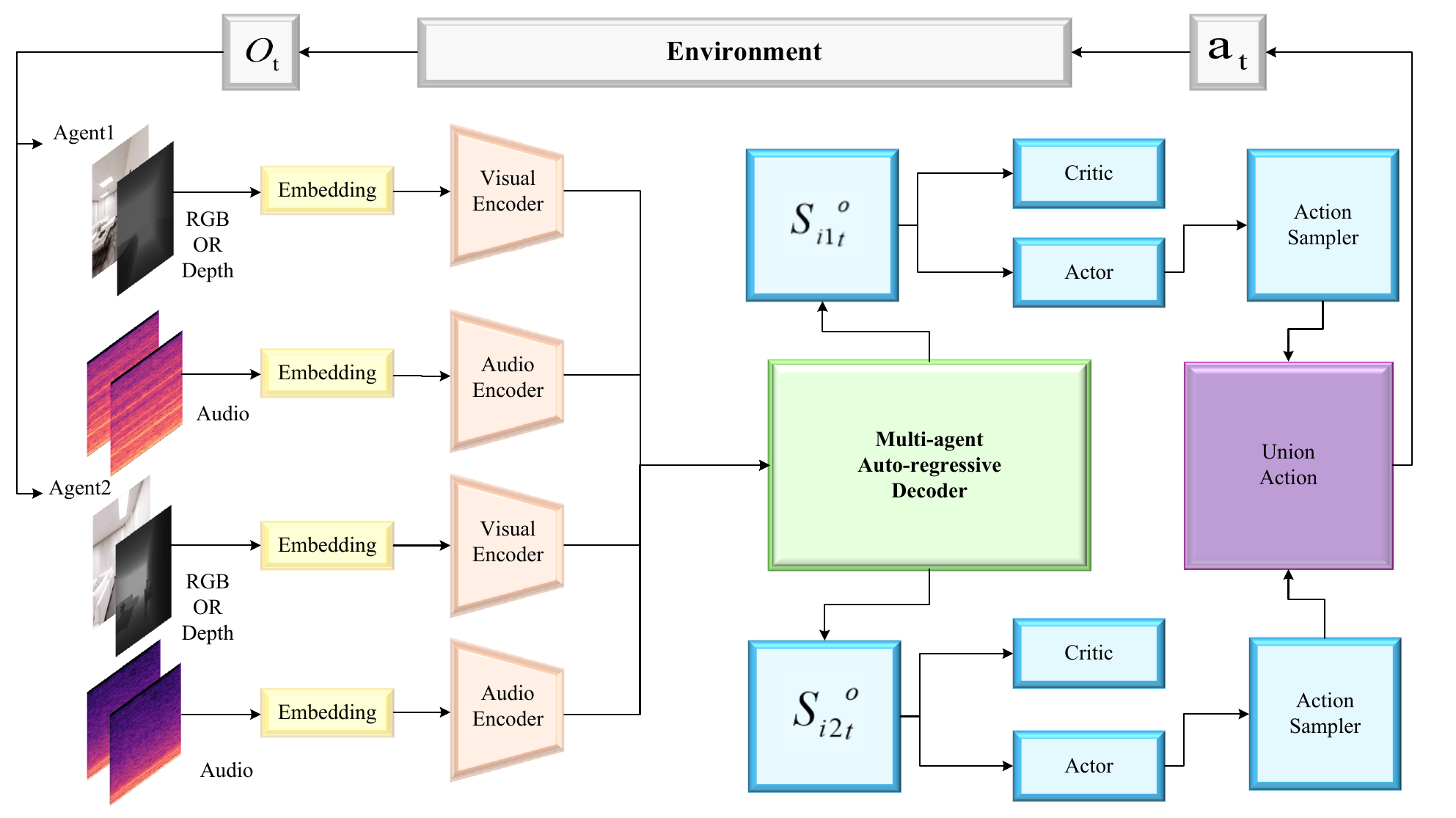}
  \caption{
   The Overview of MASTAVN: Multi-Agent Scalable Transformer Audio-Visual Navigation.
  }
  \label{fig2}
\end{figure}

\subsection{Proposed Method}

We formulate the problem as a reinforcement learning task in which two intelligent agents are required to approach a sound-emitting target in an unknown environment rapidly. Our proposed solution is called MASTAVN (Multi-Agent Scalable Transformer Audio-Visual Navigation). MASTAVN consists of four main components. Specifically, given egocentric visual and audio inputs, our model: (1) uses patch embedding to encode visual/audio inputs into visual/audio embedding vectors. (2) Each agent processes the visual/audio embeddings using modality-specific encoders to produce visual/audio information sequences. (3) The multi-modal information sequences from all agents are concatenated and passed through a shared decoder to transform them into a temporal state representation. (4) Based on the temporal state representations, each agent utilizes an actor-critic network to perform action prediction, evaluation, and policy optimization. The agents repeat this process until they reach the shared audio target in the environment. We will elaborate on each component in the following sections.

At time step $t$ during the navigation task, the $j$-th agent receives multimodal observations denoted as $O_t^{ij} = (II_t^{ij}, BI_t^{ij})$, where $II_t^{ij}$ represents the visual input and $BI_t^{ij}$ represents the audio input. The visual input is converted into a visual embedding vector, and the audio input is converted into an audio embedding vector. These embedding vectors are then processed by their corresponding encoders to generate information sequences $I_t^{ij}$ and $B_t^{ij}$. These information sequences are concatenated to form a joint state observation represented as:$C_t = \left( I_t^{i1}, B_t^{i1}, I_t^{i2}, B_t^{i2} \right).$To enable multimodal fusion and multi-agent interaction, we design a unified decoder that transforms the joint information sequence $C_t$ into temporal state representations$S_t^o = \left( (S_{i1})_t^o, (S_{i2})_t^o \right).$ for each agent.

In embodied navigation tasks, each agent's state vectors (i.e. \( s_1^{ij} \dots s_t^{ij} \)) are fed into an actor-critic network. This network serves two purposes: 1) predicting the conditional action probability distribution \( \pi^{ij}_{\theta_1^{ij}}(a_t^{ij}|s_t^{ij}) \), and 2) estimating the state value \( V^{ij}_{\theta_2^{ij}}(s_t^{ij}) \). The actor and critic are implemented using single linear layers parameterized by \( \theta_1^{ij} \) and \( \theta_2^{ij} \), respectively. For simplicity, we denote the combination of \( \theta_1^{ij} \) and \( \theta_2^{ij} \) as \( \theta^{ij} \) thereafter. The action sampler selects the actual action \( a_t^{ij} \) to execute based on \( \pi^{ij}_{\theta_1^{ij}}(a_t^{ij}|s_t^{ij}) \).

When training two agents in our navigation task, the objective is to maximize the expected discounted return \( \mathfrak{R} \), which is calculated as follows:

\begin{equation}
R = E_{\pi_{i1},\pi_{i2}}\Big[\sum_{t=1}^{T} \gamma^t (r_1(s_{t-1}^{i1}, a_t^{i1}) + r_2(s_{t-1}^{i2}, a_t^{i2}))\Big],
\end{equation}
where \( T \) is the maximum number of time steps; \(\pi_{1}, \pi_{2}\) are Policies (actors) of two agents, respectively; \( r_1(s_{t-1}^{i1}, a_t^{i1}) \) and \( r_2(s_{t-1}^{i2}, a_t^{i2}) \) represent the reward signals obtained by the agents when executing actions \( a_t^{ij} \) in state \( s_{t-1}^{ij} \), and \( \gamma \) is the discount factor that balances the weights of immediate and future rewards. The reward is calculated based on three simple rules: (1) a reward of $+10$ is given when the first agent successfully reaches the target and performs the stop action, ending the episode; (2) a reward of $+0.25$ is given when the Manhattan distance between an agent and the target decreases; (3) a time penalty of $-0.01$ is applied for each action taken by each agent.

To optimize the agents' strategies, we combine the actor loss (for updating policy parameters), critic loss (for updating value function parameters), and entropy loss of the policy distribution (to encourage exploration) to obtain the total loss function for each agent:

\begin{equation}
L_{ij} = L_{\text{actor},ij} + L_{\text{critic},ij} + L_{\text{entropy},ij},
\end{equation}
For the multi-agent system, the total loss function is obtained by summing the losses of all agents and averaging them by the number of agents:
\begin{equation}
L_{\text{total}} = \frac{1}{N} \sum_{i,j} L_{ij},
\end{equation}
Here, \( N \) represents the total number of agents. By minimizing this total loss function, the agents can continuously learn and optimize their strategies in embodied navigation tasks, thereby improving navigation's success rate and efficiency.

\begin{figure}[h] % 使用标准的浮动位置参数
  \includegraphics[width=\textwidth]{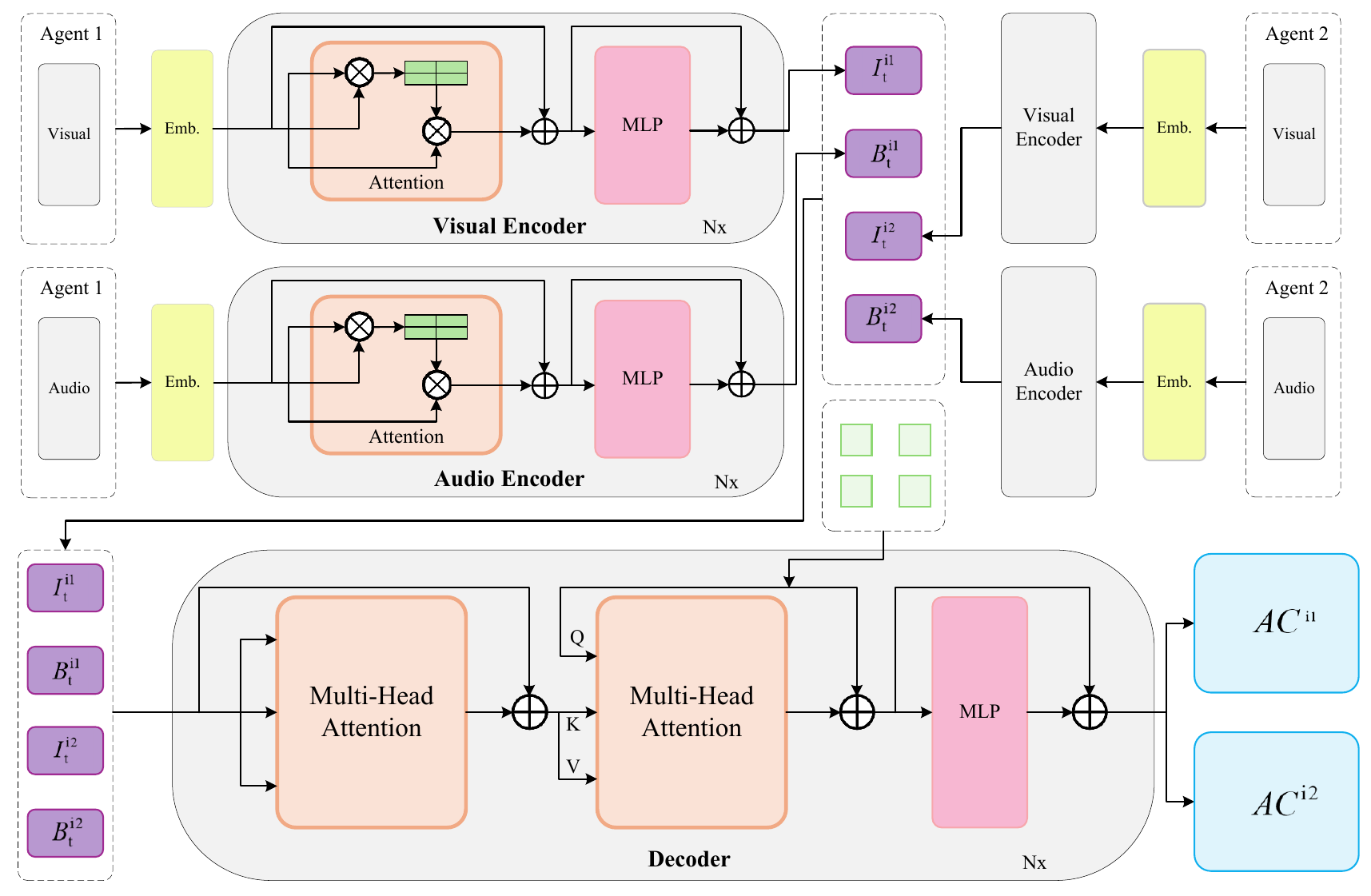}
  \caption{
  An illustrative flow diagram depicting the multi-agent scalable transformer module.
  }
  \label{fig3}
\end{figure}

\subsection{Scalable Transformer for Multi-Agent}

We propose the Multi-Agent Scalable Transformer (MAST) to enable scalable, multi-agent audio-visual navigation tasks. Inspired by the Transformer architecture, which excels at sequence modeling tasks like machine translation, our model maps the multimodal information of multiple agents into temporal state representations. Each agent then makes action decisions based on MAST's temporal state representations output. As illustrated, MAST comprises multiple encoders and a decoder. The encoders process the multimodal information from various agents, forming a joint state observation. These encoders transform the joint state observation into a unified temporal state representation for all agents.

Encoder. For each agent, visual and audio observations are input into separate visual and audio encoders. The multimodal sequences from all agents (i.e. $i_1$\ldots$i_n$) are combined into a joint state observation $C_t = \left( I_t^{i1}, B_t^{i1}\dots I_t^{in}, B_t^{in} \right)$. Each encoder block consists of a self-attention mechanism, a multi-layer perceptron (MLP), and residual connections to mitigate issues like vanishing gradients and network degradation as depth increases.

The joint state observation $C_t = \bigl(I_t^{i1}, B_t^{i1} \dots I_t^{in}, B_t^{in}\bigr)$
from all agents is fed into the decoder, which autoregressively produces a temporal state representation for each agent. The core Multi-Head Attention operation is defined as
\begin{equation}
  \mathrm{MultiHead}(T_Q, {C_t}_K, {C_t}_V) 
  = \mathrm{Concat}(\mathrm{head}_1 \dots \mathrm{head}_h)\,W^O,
\end{equation}
where each head is given by
\begin{equation}
  \mathrm{head}_i 
  = \mathrm{MASTAttention}\bigl(T_QW_i^Q,\; {C_t}_KW_i^K,\; {C_t}_VW_i^V\bigr),
\end{equation}
and
\begin{equation}
  \mathrm{MASTAttention}\bigl(T_QW_i^Q,\; {C_t}_KW_i^K,\; {C_t}_VW_i^V)
  = \mathrm{softmax}\!\bigl(\frac{T_Q\,{C_t}_K^\top}{\sqrt{d_k}}\bigr)\,{C_t}_V.
\end{equation}

We denote the decoder’s outputs by $\{S_{i1,t}^o,\; S_{i2,t}^o,\;\dots,\; S_{in,t}^o\}.$These vectors both fuse the multimodal inputs and serve as the basis for each agent’s actions. Finally, each agent uses an actor–critic network over its temporal state representation to predict actions, evaluate outcomes, and optimize its policy.

\begin{figure}[h] % 使用标准的浮动位置参数
  \includegraphics[width=\textwidth]{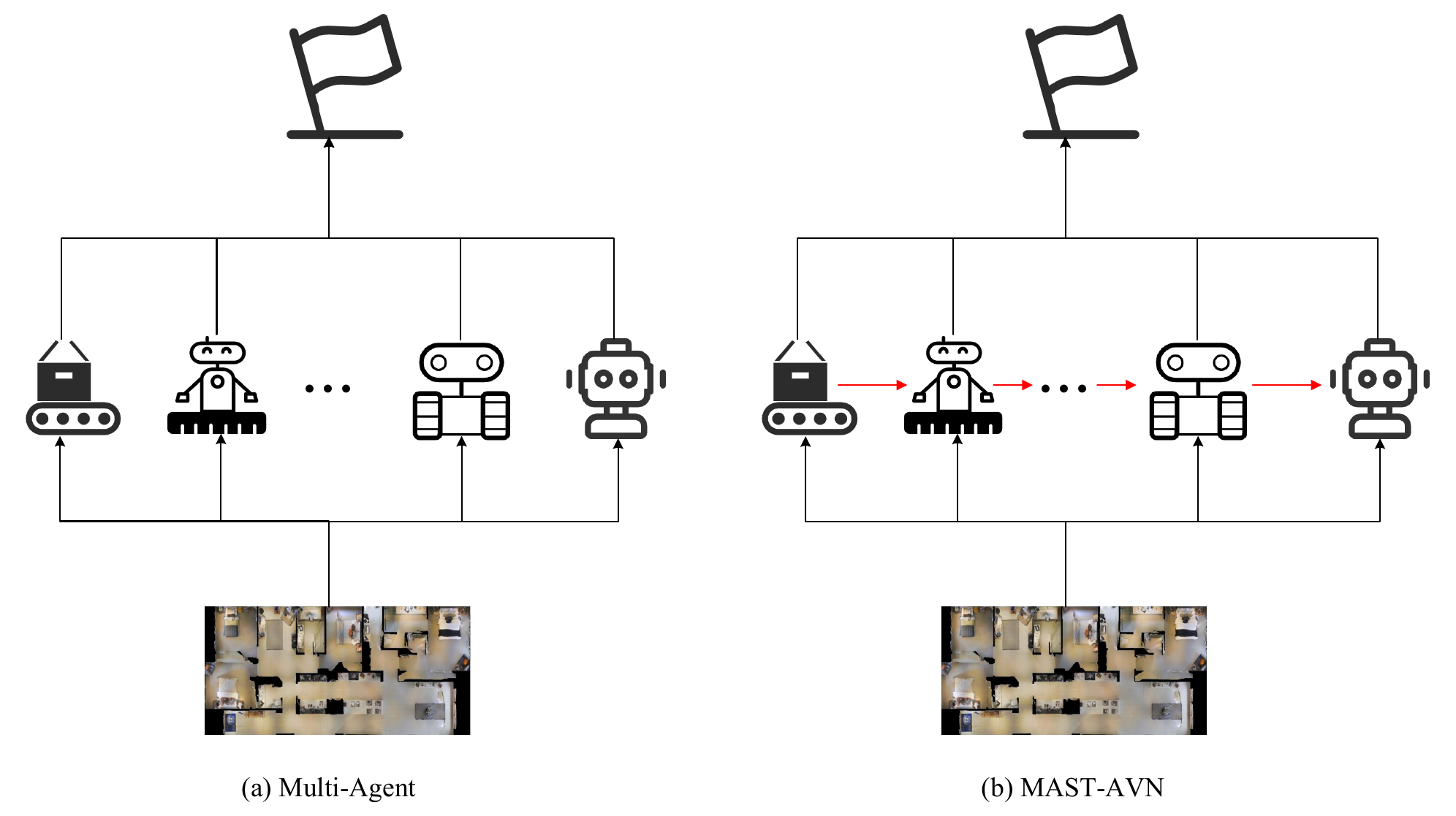}
  \caption{
  Traditional multi-agent learning paradigm (a) where all agents have no cross-modal interaction vs. multi-agent sequential interaction paradigm (b) where agents exchange modal information in sequence, and each agent incorporates the previous agent’s observation and temporal state representation, as indicated by the red arrows.
  }
  \label{fig4}
\end{figure}

\subsection{Relationship Between MAST and Sequence Modeling}

We draw inspiration from the Transformer’s sequence modeling process to enable interaction among multiple agents and propose the Multi-Agent Scalable Transformer (MAST). First, this cross-agent sequential decision framework streamlines the traditional joint policy update procedure. As illustrated in Figure~\ref{fig4}, each agent can access the observations and temporal state representations of all preceding agents in our model, making more informed and optimal decisions. Furthermore, by adjusting the number of encoder layers and the decoder’s sequence length, the model can be scaled to accommodate varying agents as needed, achieving true multi-agent scalability without treating each agent as a separate task.

\section{Experiment}
\subsection{Experimental Setup}
\subsubsection{Experimental Platform and Datasets}

In the 3D environment shown in Figure~\ref{fig1}, two intelligent agents continuously move toward a sound-emitting target within the same environment. We built a dual-agent SoundSpaces platform based on commonly used 3D environments collected from the SoundSpaces framework and the Habitat simulator. We adopt two public datasets for the audio-visual navigation task: Replica~\cite{straub2019replica} and Matterport3D~\cite{chang2017matterport3d}. Replica~\cite{straub2019replica} consists of 18 grid-based environments (with a resolution of 0.5 meters), constructed from highly accurate scans of real-world spaces such as apartments, offices, and hotels. Matterport3D~\cite{chang2017matterport3d} includes scan grids (with a resolution of 1 meter) from 85 indoor environments, encompassing realistic indoor scenes such as homes and other residential spaces. In this dual-agent platform, the omnidirectional sound emitted by the target source is convolved with the corresponding binaural Room Impulse Response (RIR). The resulting binaural environmental response simulates what the agents receive from the direction directly in front of the robot. The intelligent agents perform the audio-visual navigation task based on the real-time visual and auditory information they receive.

\subsubsection{Implementation Details}

We use the Adam optimizer (with entropy loss on the policy distribution) with a learning rate of $1 \times 10^{-4}$ to train tasks on the Replica~\cite{straub2019replica} dataset, and a learning rate of $5 \times 10^{-5}$ for the Matterport3D~\cite{chang2017matterport3d} dataset. Each episode is limited to 500 steps per agent within a scene. We train the framework for 40 million steps on Replica~\cite{straub2019replica} and 60 million steps on Matterport3D~\cite{chang2017matterport3d}.

\subsubsection{Evaluation Metrics}

We evaluate our navigation model utilizing three fundamental metrics: 1) Success Rate (SR), which quantifies the proportion of test episodes in which the agent arrives precisely at the designated audio goal location. 2) Success weighted by Path Length (SPL), which normalizes SR by the ratio of the shortest path length to the actual trajectory length, thereby rewarding both precision and navigational efficiency. 3) Success weighted by Number of Actions (SNA), which adjusts SR based on the inverse of the total number of actions executed, effectively penalizing superfluous rotations or movements that do not contribute to the agent's progress towards the goal. Specifically, in the dual-agent setting, we measure the above metrics using the data of the agent that completes the task first.

\begin{table}[h]
\centering
\caption{Comparison of Audio-Visual Navigation results on Replica.}
\label{tab:tab1}
\begin{tabular*}{\textwidth}{@{\extracolsep{\fill}} l|ccc|ccc@{}}
\toprule
\multicolumn{1}{c|}{\multirow{2}{*}{Method}} & \multicolumn{3}{c|}{Heard} & \multicolumn{3}{c}{Unheard} \\ \cmidrule(l){2-7} 
\multicolumn{1}{c|}{} & SNA $\uparrow$ & SR $\uparrow$ & SPL $\uparrow$ & SNA $\uparrow$ & SR $\uparrow$ & SPL $\uparrow$ \\ \midrule
Random Agent~\cite{chen2021waypoints} & 1.8 & 18.5 & 4.9 & 1.8 & 18.5 & 4.9 \\
Direction Follower~\cite{chen2021waypoints} & 41.1 & 72.0 & 54.7 & 8.4 & 17.2 & 11.1 \\
Frontier Waypoints~\cite{chen2021waypoints} & 35.2 & 63.9 & 44.0 & 5.1 & 14.8 & 6.5 \\
Supervised Waypoints~\cite{chen2021waypoints} & 48.5 & 88.1 & 59.1 & 10.1 & 43.1 & 14.1 \\
Gan et al.~\cite{gan2020look} & 47.9 & 83.1 & 57.6 & 5.7 & 15.7 & 7.5 \\
AV-Nav~\cite{chen2020soundspaces} & 52.7 & 94.5 & 78.2 & 16.7 & 50.9 & 34.7 \\
AV-WaN~\cite{chen2021waypoints} & 70.7 & 98.7 & 86.6 & 27.1 & 52.8 & 34.7 \\
ORAN~\cite{chen2023omnidirectional} & 70.1 & 96.7 & 84.2 & 36.5 & 60.9 & 46.7 \\
MAST-AVN (ours) & \textbf{84.6} & \textbf{99.9} & \textbf{99.1} & \textbf{53.8} & \textbf{72.4} & \textbf{59.3} \\ \bottomrule
\end{tabular*}
\end{table}

\begin{table}[h]
\centering
\caption{Comparison of Audio-Visual Navigation results on Matterport3D}
\label{tab:tab2}
\begin{tabular*}{\textwidth}{@{\extracolsep{\fill}} l|ccc|ccc@{}}
\toprule
\multicolumn{1}{c|}{\multirow{2}{*}{Method}} & \multicolumn{3}{c|}{Heard} & \multicolumn{3}{c}{Unheard} \\ \cmidrule(l){2-7} 
\multicolumn{1}{c|}{} & SNA $\uparrow$ & SR $\uparrow$ & SPL $\uparrow$ & SNA $\uparrow$ & SR $\uparrow$ & SPL $\uparrow$ \\ \midrule
Random Agent~\cite{chen2021waypoints} & 0.8 & 9.1 & 2.1 & 0.8 & 9.1 & 2.1 \\
Direction Follower~\cite{chen2021waypoints} & 23.8 & 41.2 & 32.3 & 10.7 & 18.0 & 13.9 \\
Frontier Waypoints~\cite{chen2021waypoints} & 22.2 & 42.8 & 30.6 & 8.1 & 16.4 & 10.9 \\
Supervised Waypoints~\cite{chen2021waypoints} & 16.2 & 36.2 & 21.0 & 2.9 & 8.8 & 4.1 \\
Gan et al.~\cite{gan2020look} & 17.1 & 37.9 & 22.8 & 3.6 & 10.2 & 5.0 \\
AV-Nav~\cite{chen2020soundspaces} & 32.6 & 71.3 & 55.1 & 12.8 & 40.1 & 25.9 \\
AV-WaN~\cite{chen2021waypoints} & 54.8 & 93.6 & 72.3 & 30.6 & 56.7 & 40.9 \\
ORAN~\cite{chen2023omnidirectional} & 57.7 & 93.5 & 73.7 & 35.3 & 59.4 & 50.8 \\
MAST-AVN (ours) & \textbf{73.8} & \textbf{98.9} & \textbf{91.6} & \textbf{48.5} & \textbf{73.2} & \textbf{64.7} \\ \bottomrule
\end{tabular*}
\end{table}

\subsection{Performance Comparison}

This work introduces the Multi-Agent Scalable Transformer (MAST), marking the first scalable multi-agent environment for audio-visual embodied navigation in 3D spaces. To demonstrate the effectiveness of MAST, we compare it with prior single-agent audio-visual navigation baselines, including those by Chen et al.~\cite{chen2020soundspaces,chen2021waypoints}, Gan et al.~\cite{gan2020look}, and Chen et al.~\cite{chen2023omnidirectional}.

On the Replica~\cite{straub2019replica} dataset, as shown in Table~\ref{tab:tab1}, quantitative comparisons reveal that our MAST-AVN model outperforms previous single-agent methods across all metrics, both in `Heard' and `Unheard' scenarios. Specifically, MAST-AVN surpasses ORAN~\cite{chen2023omnidirectional} by 14.5 in SNA\texttt{@}Heard, 3.2 in SR\texttt{@}Heard, 14.9 in SPL\texttt{@}Heard, 7.3 in SNA\texttt{@}Unheard, 11.5 in SR\texttt{@}Unheard, and 12.6 in SPL\texttt{@}Unheard.  Furthermore, our model achieves significant performance improvements compared to AV-WaN~\cite{chen2021waypoints}, the current state-of-the-art waypoint-based baseline.  These results underscore the advantages of dual-agent systems in audio-visual navigation and highlight the superiority of our approach over advanced baselines.

Additionally, Table~\ref{tab:tab2} showcases notable enhancements on the Matterport3D~\cite{chang2017matterport3d} benchmark. Compared to AV-WaN~\cite{chen2021waypoints}, our model achieves gains of 19.0 in SNA\texttt{@}Heard, 5.3 in SR\texttt{@}Heard, 19.3 in SPL\texttt{@}Heard, 17.9 in SNA\texttt{@}Unheard, 16.5 in SR\texttt{@}Unheard, and 23.8 in SPL\texttt{@}Unheard. We also observe strong performance against AV-Nav~\cite{chen2020soundspaces}, a deep reinforcement learning network based on separate audio-visual encoders.  These findings demonstrate the efficacy of MAST in multimodal information fusion and implicit multi-agent interaction.

\begin{table}[h]
\centering
\caption{Comparison of different Dual-Agent Navigation methods on Replica.}
\label{tab:tab3}
\begin{tabular*}{\textwidth}{@{\extracolsep{\fill}} l|ccc|ccc@{}}
\toprule
\multicolumn{1}{c|}{\multirow{2}{*}{Method}} & \multicolumn{3}{c|}{Heard} & \multicolumn{3}{c}{Unheard} \\ \cmidrule(l){2-7} 
\multicolumn{1}{c|}{} & SNA $\uparrow$ & SR $\uparrow$ & SPL $\uparrow$ & SNA $\uparrow$ & SR $\uparrow$ & SPL $\uparrow$ \\ \midrule
Random Agent (Two agent) & 2.3 & 19.6 & 5.6 & 2.3 & 19.6 & 5.6 \\
AV-Nav (Two agent) & 60.6 & 95.2 & 82.4 & 33.7 & 59.9 & 45.7 \\
MAST-AVN (ours) & \textbf{84.6} & \textbf{99.9} & \textbf{99.1} & \textbf{53.8} & \textbf{72.4} & \textbf{59.3} \\ \bottomrule
\end{tabular*}
\end{table}

\begin{table}[h]
\centering
\caption{Comparison of different Dual-Agent Navigation methods on Matterport3D}
\label{tab:tab4}
\begin{tabular*}{\textwidth}{@{\extracolsep{\fill}} l|ccc|ccc@{}}
\toprule
\multicolumn{1}{c|}{\multirow{2}{*}{Method}} & \multicolumn{3}{c|}{Heard} & \multicolumn{3}{c}{Unheard} \\ \cmidrule(l){2-7} 
\multicolumn{1}{c|}{} & SNA $\uparrow$ & SR $\uparrow$ & SPL $\uparrow$ & SNA $\uparrow$ & SR $\uparrow$ & SPL $\uparrow$ \\ \midrule
Random Agent (Two agent) & 1.3 & 10.2 & 2.9 & 1.3 & 10.2 & 2.9 \\
AV-Nav (Two agent) & 44.6 & 76.3 & 63.1 & 26.8 & 50.1 & 37.9 \\
MAST-AVN (ours) & \textbf{73.8} & \textbf{98.9} & \textbf{91.6} & \textbf{48.5} & \textbf{73.2} & \textbf{64.7} \\ \bottomrule
\end{tabular*}
\end{table}

\subsection{Dual-Agent Comparison}
% In this section, given that our audio-visual navigation task is designed to control two agents simultaneously, we made only minimal modifications to the original SoundSpaces baseline (AV-Nav): we simply placed both agents in the same environment to perform the navigation task, without introducing any additional interaction mechanisms between them. As shown in Tables~\ref{tab:tab3} and~\ref{tab:tab4}, across both the Replica and Matterport3D datasets, our proposed MAST substantially outperforms the original AV-Nav across all evaluation metrics. Moreover, as illustrated in Figure 5 with side-by-side trajectory visualizations, under our multi-agent setup both agents navigate more accurately and efficiently to the audio target, further confirming the advantages and state-of-the-art performance of MAST in multi-agent audio-visual navigation tasks.

\begin{figure}[h] % 使用标准的浮动位置参数
  \includegraphics[width=\textwidth]{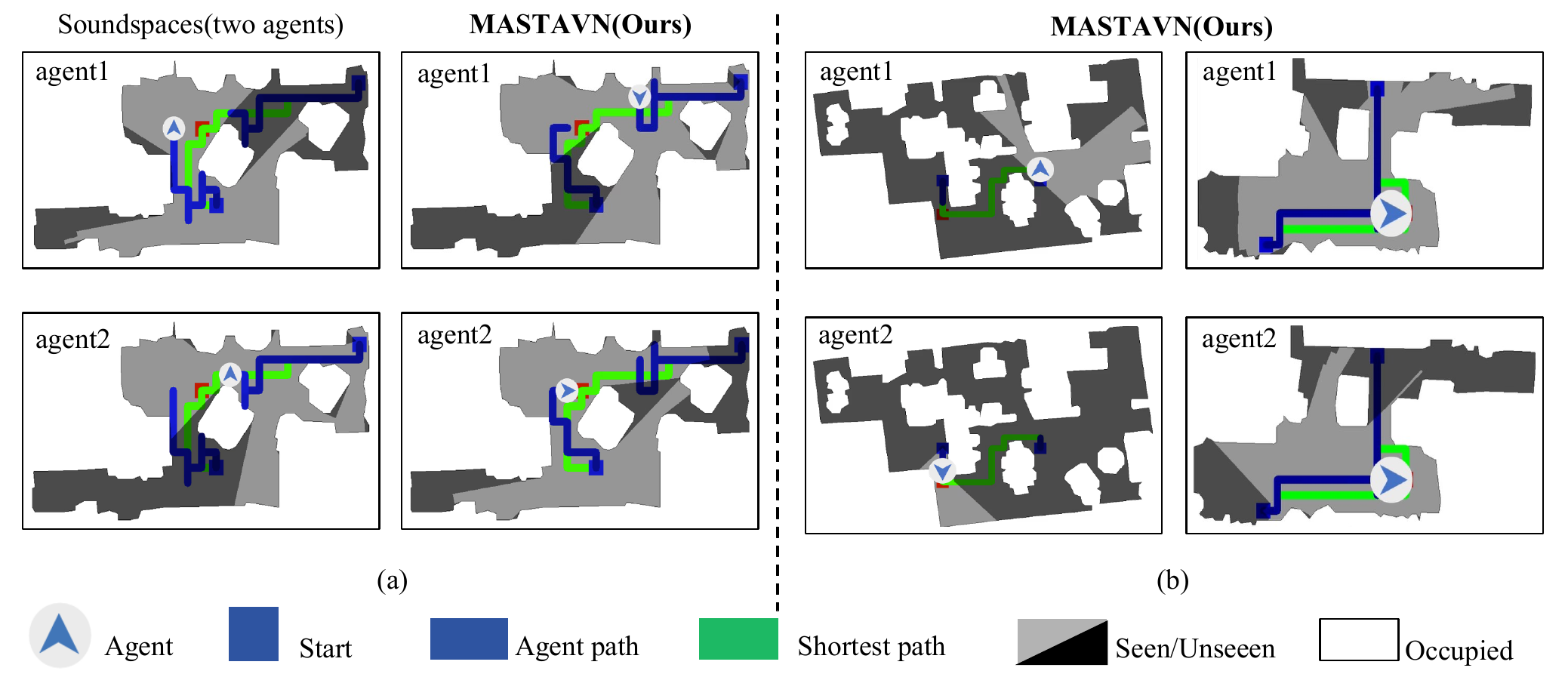}
  \caption{
  Figure~\ref{fig5}(a) offers an intuitive comparison of the trajectories of dual agents across distinct methods. In contrast, Figure~\ref{fig5}(b) underscores the critical role of collaboration between the two agents within the environment.
  }
  \label{fig5}
\end{figure}

In this section, given that our audio-visual navigation task is designed to control two agents simultaneously, we made only minimal modifications to the original SoundSpaces~\cite{chen2020soundspaces} baseline (AV-Nav): we placed both agents in the same environment to perform the navigation task, without introducing any additional interaction mechanisms between them. As shown in Tables~\ref{tab:tab3} and~\ref{tab:tab4}, across both the Replica~\cite{straub2019replica} and Matterport3D~\cite{chang2017matterport3d} datasets, our proposed MAST outperforms the original AV-Nav across all evaluation metrics. Moreover, as illustrated in Figure~\ref{fig5} with side-by-side trajectory visualizations, both agents navigate more accurately and efficiently to the audio target under our multi-agent setup, further confirming the advantages and state-of-the-art performance of MAST in multi-agent audio-visual navigation tasks.

\begin{table}[h]
\centering
\caption{Ablation study on the proposed model on Replica.}
\label{tab:tab5}
\begin{tabular*}{\textwidth}{@{\extracolsep{\fill}} l|ccc|ccc@{}}
\toprule
\multirow{2}{*}{Ablation} & \multicolumn{3}{c|}{Heard} & \multicolumn{3}{c}{Unheard} \\ \cmidrule(l){2-7}
                          & SNA($\uparrow$) & SR($\uparrow$) & SPL($\uparrow$) & SNA($\uparrow$) & SR($\uparrow$) & SPL($\uparrow$) \\ \midrule
MAST-AVN(Ours)            & \textbf{84.6}   & \textbf{99.9}  & \textbf{99.1}   & \textbf{53.8}   & \textbf{72.4}  & \textbf{59.3}   \\
w/o EN                    & 47.6            & 76.2           & 65.8            & 24.7            & 46.3           & 34.1            \\
w/o DE                    & 33.4            & 70.9           & 55.2            & 12.6            & 39.5           & 25.4            \\ \bottomrule
\end{tabular*}
\end{table}

\begin{table}[h]
\centering
\caption{Ablation study on the proposed model on Matterport3D}
\label{tab:tab6}
\begin{tabular*}{\textwidth}{@{\extracolsep{\fill}} l|ccc|ccc@{}}
\toprule
\multirow{2}{*}{Ablation} & \multicolumn{3}{c|}{Heard} & \multicolumn{3}{c}{Unheard} \\ \cmidrule(l){2-7}
                          & SNA($\uparrow$) & SR($\uparrow$) & SPL($\uparrow$) & SNA($\uparrow$) & SR($\uparrow$) & SPL($\uparrow$) \\ \midrule
MAST-AVN(Ours)            & \textbf{73.8}   & \textbf{98.9}  & \textbf{91.6}   & \textbf{48.5}   & \textbf{73.2}  & \textbf{64.7}   \\
w/o EN                    & 42.6            & 73.3           & 62.2            & 28.9            & 51.3           & 37.4            \\
w/o DE                    & 27.1            & 65.4           & 50.3            & 15.7            & 43.7           & 31.7            \\ \bottomrule
\end{tabular*}
\end{table}

\subsection{Ablation Study}

We conduct ablation experiments on the Replica~\cite{straub2019replica} and Matterport3D~\cite{chang2017matterport3d} datasets. Table~\ref{tab:tab5} and Table~\ref{tab:tab6} show the results under different ablation settings.

\subsubsection{w/o EN}

\begin{figure}[h] % 使用标准的浮动位置参数
  \includegraphics[width=\textwidth]{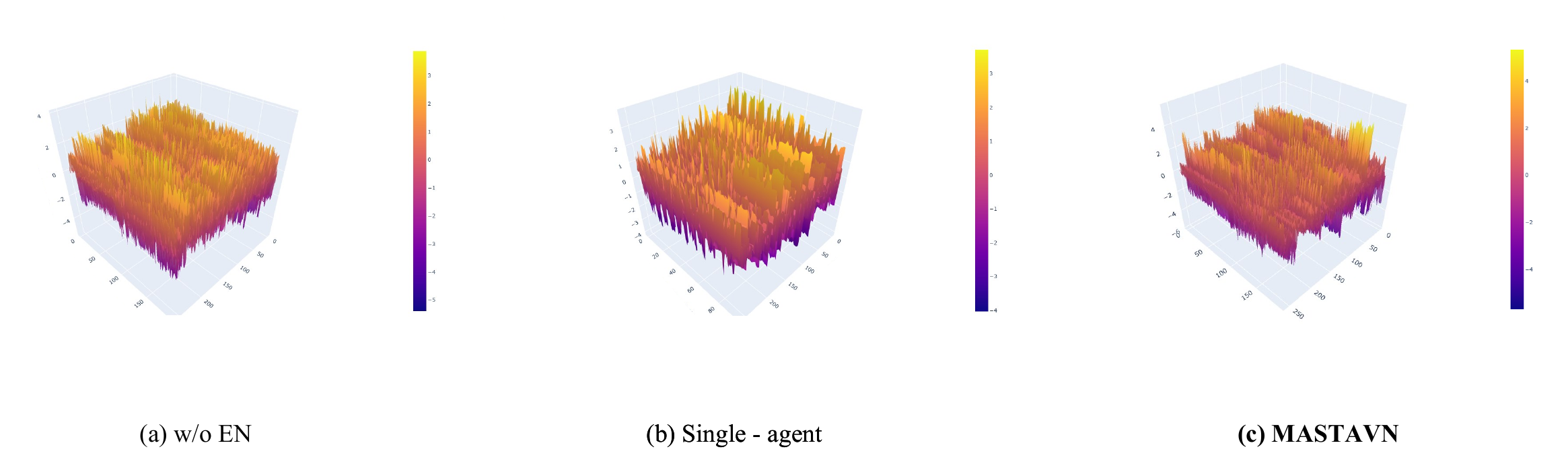}
  \caption{
  Comparison of 3D Attention Weight Visualizations Across Multiple Scenarios.
  }
  \label{fig6}
\end{figure}

Removing the EN module means removing the visual and audio encoders from the model. In this case, the multi-modal embedding vectors from multiple agents are directly concatenated and fed into the decoder without being processed by the encoders. The multi-head self-attention layers in the visual and audio encoders compute the similarity between elements in the sequence and assign different weights to them. This mechanism enables fine-grained control over multi-modal information and performs pre-processing for the decoding stage. Without the EN module, the model struggles to accurately capture the key aspects of multi-modal details, leading to a significant drop in navigation performance. Figure~\ref{fig6} (a) illustrates the imprecise attention weights of multiple agents without the EN module, resulting in poor navigation. Figure~\ref{fig6} (b) shows the attention weights of a single agent, where the lack of collaboration limits navigation efficiency. Figure~\ref{fig6} (c) demonstrates the effective multimodal fusion and agent coordination under the MASTAVN model, leading to significantly improved navigation performance.

\subsubsection{w/o DE}

Removing the DE module implies that the model no longer performs decoding; instead, it uses MLPs to map the joint state observation directly into temporal state representations. The DE module is crucial in facilitating implicit communication between agents, enabling deep interaction across agents and modalities. Without the DE module, the model fails to effectively decode and transform the joint observations, making it difficult for agents to share and utilize information. This, in turn, results in suboptimal decision-making and degraded navigation performance.

These ablation results demonstrate that each module is essential. They enhance the model’s ability to make accurate path selection decisions.

\section{Conclusion}

In this paper, we are the first to realize a scalable multi-agent task in the field of audio-visual navigation, and we introduce MASTAVN (Multi-Agent Scalable Transformer Audio-Visual Navigation) to achieve more efficient audio-visual navigation in multi-agent settings. This approach fuses multimodal information and enables deep interaction among agents by treating the joint multi-agent multimodal input as a sequence for modeling. It transforms the complex joint policy optimization problem into a sequential state representation process that an autoregressive model can handle. Moreover, the model is truly scalable: by expanding the number of encoder layers and the dimensionality of the decoder, it can be extended from two agents to many.

Furthermore, our experiments on the Replica and Matterport3D datasets demonstrate that our method outperforms traditional approaches at executing multi-agent tasks. By employing a scalable, sequence-modeling transformer across multiple agents, our model achieves not only higher SR (Success Rate) but also surpasses the baselines on SPL (Success weighted by Path Length) and SNA (Success weighted by Number of Agents). Together, these metrics confirm the effectiveness of our model in tasks involving multiple intelligent agents.

This work opens new directions for future research. One potential avenue is exploring novel forms of interaction among agents in audio-visual navigation environments, such as explicit communication, to optimize multi-agent task efficiency. We can also investigate different task types and assign specialized agent configurations to match each task’s requirements. By demonstrating our model’s effectiveness and significance in multi-agent interaction scenarios, we contribute to the rapidly evolving field of embodied artificial intelligence.

\section*{Acknowledgements}

This research was financially supported by the National Natural Science Foundation of China (Grants Nos. 62463029, 62472368, and 62303259) and the Natural Science Foundation of Tianjin (Grant No. 24JCQNJC00910).
%

\bibliographystyle{splncs04}

% \bibliographystyle{IEEEtran}   % IEEE会议默认使用的参考文献样式
\bibliography{mybib}

\end{document}